\documentclass{article}
\usepackage{hyperref}
\usepackage{graphicx}
\usepackage{arxiv}
\usepackage{tikz}
\usetikzlibrary{matrix}

\usepackage{booktabs} 
\usepackage{algorithm}
\usepackage{algpseudocode}
\usepackage{xcolor}
\usepackage{float}
\usepackage{natbib}

\title{Preliminary Tests of the Anticipatory Classifier System with Hindsight Experience Replay}

\author{
    Olgierd Unold \href{https://orcid.org/0000-0003-4722-176X}{\includegraphics[scale=0.06]{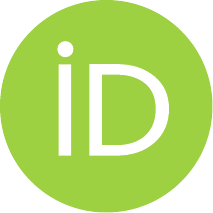}}\thanks{\url{https://ounold.github.io/}} \\
    Department of Computer Engineering \\
    Wroclaw University of Science and Technology \\
    Wyb. Wyspiańskiego 27, 50-370 Wrocław, Poland \\
    \texttt{olgierd.unold@pwr.edu.pl} 
    \and
    \textbf{Stanisław Franczyk} \\
    Department of Computer Engineering \\
    Wroclaw University of Science and Technology \\
    Wyb. Wyspiańskiego 27, 50-370 Wrocław, Poland \\
    \texttt{}
}

\begin{document}
\maketitle

\begin{abstract}
This paper introduces ACS2HER, a novel integration of the Anticipatory Classifier System (ACS2) with the Hindsight Experience Replay (HER) mechanism. While ACS2 is highly effective at building cognitive maps through latent learning, its performance often stagnates in environments characterized by sparse rewards. We propose a specific architectural variant that triggers hindsight learning when the agent fails to reach its primary goal, re-labeling visited states as virtual goals to densify the learning signal. The proposed model was evaluated on two benchmarks: the deterministic \texttt{Maze 6} and the stochastic \texttt{FrozenLake}. The results demonstrate that ACS2HER significantly accelerates knowledge acquisition and environmental mastery compared to the standard ACS2. However, this efficiency gain is accompanied by increased computational overhead and a substantial expansion in classifier numerosity. This work provides the first analysis of combining anticipatory mechanisms with retrospective goal-relabeling in Learning Classifier Systems.
\end{abstract}


\maketitle

\label{sec:introduction}
The Learning Classifier Systems (LCS) concept was introduced by John Holland, one of the inventors of computer algorithms that mimic genetic evolution, in the 1970s \cite{holland1978cognitive}. The term LCS describes the family of rule-based machine learning algorithms that utilize the paradigm of Reinforcement Learning (RL) to model complex adaptive systems \cite{Urbanowicz:2017:ILC:3154527}. The Accuracy-based Classifier System (XCS) is probably the most well-known and investigated instance of LCS, both analytically \cite{patzel2019survey} and empirically \cite{hansmeier2021experimental}. 

Anticipatory Classifier System (ACS) is a variant of the Learning Classifier System extending the classical human-interpretable rule-based model with the psychological theory of anticipations \cite{IntroToACS,butz2002anticipatory}. Unlike purely reactive systems, ACS builds an internal model of the environment’s dynamics, allowing it to anticipate the sensory consequences of its actions. Every situation is accompanied by consequences after performing each possible behavior. The most investigated representative of Anticipatory Learning Classifier Systems (ALCS) is ACS2 \cite{AlgorithmicDescriptionOfACS2}. 

Recently, \textbf{Unold et al. extended the ACS2 Classifier System with a replay buffer called Experience Replay (ER)} \cite{unold2022preliminary}. ER is an integral part of the Deep Q-Network (DQN) family and is mainly used to stabilize the neural network's training and increase learning efficiency. Empirical evidence indicates that this hybrid approach significantly enhances both the learning efficiency and the generality of the rules discovered relative to the original ACS2 implementation.

Despite its success, ACS2 faces a significant hurdle: the sparse reward problem. In many real-world scenarios, an agent only receives a meaningful signal when reaching a terminal goal state. If exploration fails to encounter this goal, the Reinforcement Learning component receives no feedback, and the Anticipatory Learning Process (ALP) remains confined to the narrow subset of visited states. To address this, we propose the integration of Hindsight Experience Replay. HER allows an agent to learn from failure by "pretending" that a state it accidentally reached was actually its goal, thereby turning every trajectory into a dense learning opportunity.

This paper aims to \textbf{propose the ACS2HER model} and test it over deterministic and stochastic multi-step problems. The main scientific contributions of this research include:
\begin{itemize}
    \item \textbf{A New Algorithmic Framework}: The proposal of ACS2HER, which integrates a Replay Memory buffer ($RM$) and a goal-relabeling strategy designed to activate specifically upon trial failure.
    \item \textbf{Comparative Study}: A rigorous analysis of ACS2, ACS2ER, and ACS2HER over RL problems taken from OpenAI gym \cite{2016arXiv160601540B,kozlowski2018integrating}, specifically the \texttt{Maze6} and \texttt{FrozenLake} environments.
    \item \textbf{Knowledge vs. Complexity Analysis}: A first insight into the performance differences between examined \textit{m}-sized batches of memorized experiences and the $k$ hindsight factor, highlighting the trade-offs in learning efficiency and classifier numerosity.
    \item \textbf{Experimental Evidence}: Preliminary results indicating that ACS2 expanded by Hindsight Experience Replay can significantly increase the rate of knowledge acquisition, though at the cost of increased computational time and population size.
\end{itemize}

The remainder of the paper is organized as follows. Section \ref{sec:background} provides an overview of related work concerning memory and experience replay in the LCS domain and establishes the theoretical foundations of Anticipatory Learning Classifier Systems and Hindsight Experience Replay. The proposed ACS2HER architecture, including its goal-relabeling logic and the update cycle, is detailed in Section \ref{sec:acs2-her}. Section \ref{sec:experiments} describes the experimental methodology and provides a detailed analysis of the results obtained in the \texttt{Maze~6} and \texttt{FrozenLake} environments. Section \ref{sec:conclusions} summarizes the findings and describes prospective directions for future research. Finally, a critical evaluation of the model's constraints and computational overhead is presented in Section \ref{sec:limit}.

\section{Background}
\label{sec:background}

\subsection{Anticipatory Learning Classifier Systems (ACS2)}
\label{back:acs2}
The Anticipatory Learning Classifier System (ACS2) constitutes a specialized extension of the Learning Classifier Systems (LCS) paradigm, initially formulated by Holland \cite{holland1978cognitive}. In contrast to conventional LCS variants—such as the accuracy-driven XCS, which emphasize stimulus-response mappings—ACS2 \cite{butz2002anticipatory} operationalizes Hoffmann's theory of anticipative behavioral control \cite{hoffmann}, thereby integrating predictive foresight into classifier evolution.

Knowledge in ACS2 is represented as a population $[P]$ of classifiers, each utilizing a structure $Condition-Action-Effect$ ($C-A-E$). This triplet allows the agent not only to suggest an action for a given state but also to predict the subsequent state of the environment. The learning architecture is driven by three primary mechanisms:
\begin{itemize}
    \item \textbf{Anticipatory Learning Process (ALP):} The core mechanism that evolves the population by comparing anticipated effects ($E_{ant}$) with real environmental feedback ($E_{real}$). Rules are specialized via a "covering" process or generalized to increase their environmental coverage.
    \item \textbf{Reinforcement Learning (RL):} A weight-updating procedure, typically based on Q-learning, that associates reward expectations with specific classifiers.
    \item \textbf{Genetic Algorithm (GA):} An evolutionary process that explores the rule space to discover more efficient and general representations of environmental dynamics.
\end{itemize}
The model-based nature of ACS2 enables \textit{latent learning}, allowing the agent to construct a cognitive map of the environment even in the absence of external rewards. Furthermore, the IF-THEN nature of the rules ensures high model interpretability, aligning with the objectives of Explainable Artificial Intelligence (XAI).

\begin{algorithm}
  \caption{ACS2}\label{list:acs2}
  \begin{algorithmic}
  \Require $trials $ 
  \Require $do\_ga \gets True$ or $False$ 
  \State $current\_trial \gets 1$
  \State $t \gets 0$
  \While{$current\_trail \leq trials$}  
    \State $done \gets False$ 
    \State $s_t \gets env.reset()$ 
    \State $[A]_{t-1} \gets \emptyset$ 
    \While{$done$ is $False$} 
      \State $[M] \subseteq [P]$ for $s_t$ 
      \If{is NOT first step of episode} 
        \State Run ALP on $[A]_{t-1}$ considering $[M], s_{t-1}, a_{exec}, s_t, t$ 
        \State Run RL on $[A]_{t-1}$ considering $r_{t-1}, [M]$ 
        \If{$do\_ga$ and $t - \frac{\Sigma_{cl\in[A]t-1}cl.num * cl.tga}{\Sigma_{cl\in[A]t-1}cl.num} >  \Theta_{GA}$} 
          \State Run GA on $[A]_{t-1}$ considering $[P], [M], s_t, t$ 
        \EndIf
      \EndIf

      \If{$U[0,1] < \epsilon$} 
        \State $a_{exec} \gets rand(A)$ 
      \Else
        \State $a_{exec} \gets best(A)$ 
      \EndIf
      \State $[A]_t \subseteq [M]$ for $a_{exec}$ 
      \State $s_{t-1} \gets s_t$
      \State $s_t, r_{t-1}, done \gets env.step(a_{exec})$ 
      \State $[A]_{t-1} \gets [A]_t$

      \If{$done$ is $True$} 
        \State Run ALP on  $[A]_{t-1}$ considering $s_{t-1}, a_{exec}, s_t, t$ 
        \State Run RL on $[A]_{t-1}$ considering $r_{t-1}$ 
        \If{$do\_ga$ and $t - \frac{\Sigma_{cl\in[A]t-1}cl.num * cl.tga}{\Sigma_{cl\in[A]t-1}cl.num} >  \Theta_{GA}$} 
          \State Run GA on $[A]_{t-1}$ considering $[P], s_t, t$ 
        \EndIf
      \EndIf

      \State $t \gets t + 1$
    \EndWhile
    \State $current\_trial \gets current\_trial + 1$
  \EndWhile
  \end{algorithmic}
\end{algorithm}

The operational dynamics of ACS2 are formalized in Algorithm \ref{list:acs2}. The learning procedure is organized into a predefined number of episodes, or trials, each initiated by an environmental reset to a starting configuration. Within each discrete time step, the system generates a match set $[M]$ consisting of all classifiers in the population $[P]$ whose condition parts are satisfied by the current sensory input. 

An action $a_{exec}$ is subsequently selected following an $\epsilon$-greedy policy: with probability $1-\epsilon$, the agent selects the action associated with the highest predicted payoff (exploitation), while with probability $\epsilon$, a random action is chosen to facilitate exploration. Once the action is executed, an action set $[A]$ is formed, containing all classifiers from $[M]$ that proposed $a_{exec}$. 

The learning phase then proceeds through two concurrent mechanisms: the Anticipatory Learning Process (ALP), which refines the structural accuracy of the $C-A-E$ rules, and Reinforcement Learning (RL), which updates the payoff expectations based on the observed reward $r$. In multi-step environments, this iterative cycle continues until a terminal state is reached; in single-step tasks, the trial concludes immediately following the initial update. Furthermore, the Genetic Algorithm (GA) may be invoked with a probability $\Theta_{GA}$ to foster the discovery of novel, generalized rule structures by recombining high-fitness individuals within $[A]$.

\subsection{Experience Replay (ER)}
\label{back:er}

\begin{algorithm}
  \caption{ACS2ER}\label{list:acs2er}
  \begin{algorithmic}
  \Require $trials \gets $ 
  \Require $do\_ga \gets True$ or $False$ 
  \Require $N$ 
  \Require $m$ 
  \Require $N\_warmup$ 
  \State $current\_trial \gets 1$
  \State $t \gets 0$
  \State $RM \gets \emptyset$ 
  \While{$current\_trail \leq trials$}  
    \State $done \gets False$ 
    \State $s_t \gets env.reset()$ 
    \State $[A]_{t-1} \gets \emptyset$ 
    \While{$done$ is $False$} 
      \State $[M] \subseteq [P]$ for $s_t$ 
      \If{$U[0,1] < \epsilon$} 
        \State $a_{exec} \gets rand(A)$ 
      \Else
        \State $a_{exec} \gets best(A)$ 
      \EndIf
      \State $s_{t-1} \gets s_t$
      \State $s_t, r_{t-1}, done \gets env.step(a_{exec})$ 
      \State $a_{t-1} \gets a_{exec}$

      \If{size of $RM \geq N$} 
        \State Drop the oldest experience $RM[0]$ 
      \EndIf

      \State $RM := RM \cup \{(s_{t-1}, a_{t-1}, r_{t-1}, s_t)\}$ 

      \If{size of $RM \geq N\_warmup$} 
        \State $SM$ of size $m \subseteq RM$ 
        \ForAll{$sm$ in $SM$}  
          \State $s, a, r, s' \gets sm$
          \State $[M] \subseteq [P]$ for $s$ 
          \State $[A] \subseteq [M]$ for $a$ 
          \State $[M]' \subseteq [P]$ for $s'$ 

          \State Run ALP on $[A]$ considering $[M]', s, a, s', t$ 
          \State Run RL on $[A]$ considering $r, [M]'$ 
          \If{$do\_ga$ and $t - \frac{\Sigma_{cl\in[A]}cl.num * cl.tga}{\Sigma_{cl\in[A]}cl.num} >  \Theta_{GA}$} 
            \State Run GA on $[A]$ considering $[P], [M]', s', t$ 
          \EndIf
        \EndFor
      \EndIf

      \State $t \gets t + 1$
    \EndWhile
    \State $current\_trial \gets current\_trial + 1$
  \EndWhile
  \end{algorithmic}
  \end{algorithm}

In classical Reinforcement Learning, agents typically learn from sequential transitions $(s, a, r, s')$, where each experience is discarded after a single update. Experience Replay \cite{lin1992self} decouples this temporal dependency by introducing a Replay Memory ($RM$) buffer. During the training phase, the agent stores transitions in $RM$ and periodically samples mini-batches to perform updates. 

This mechanism provides several critical advantages for the learning process:
\begin{enumerate}
    \item \textbf{Sample Efficiency:} Each experience can be reused multiple times for learning updates.
    \item \textbf{Stability:} Random sampling from $RM$ breaks the correlation between consecutive samples, preventing the model from over-fitting to the most recent trajectory—a feature that significantly stabilizes training in deep RL architectures \cite{mnih2013playing}.
    \item \textbf{Data Distribution Smoothing:} It allows the agent to re-learn from rare but high-utility transitions that might otherwise be forgotten.
\end{enumerate}

To the best of our knowledge, the integration of Experience Replay or analogous transition-memory mechanisms within the Learning Classifier System (LCS) paradigm remains a relatively nascent area of investigation. Recent contributions by Stein et al. \cite{stein2020xcs} formalized the implementation of ER within the XCS framework (XCS-ER), demonstrating its efficacy in complex domains, such as automated test case prioritization \cite{rosenbauer2020xcs, rosenbauer2020xcsf}. This followed foundational research \cite{stein2018interpolation} in which historical input-reward pairs were utilized to facilitate classifier prediction modeling using interpolation techniques.

\textbf{\cite{unold2022preliminary} introduces ACS2ER}, representing the first formal integration of an \textbf{Experience Replay} mechanism within the Anticipatory Classifier System framework. By decoupling environmental interaction from the learning process through a Replay Memory (RM) buffer, the architecture allows for the systematic re-sampling and re-utilization of historical transitions.
Empirical evaluations across both single- and multi-step discrete environments demonstrate that this synergy significantly enhances \textbf{learning efficiency} and the \textbf{generality of the discovered rules} compared to the standard online ACS2 model.
The operational flow of ACS2ER is structured as follows (see Algorithm \ref{list:acs2er}). The system initializes an empty Replay Memory $RM$ with a maximum capacity $N$. During each trial, the agent interacts with the environment by generating a match set $[M]$ and selecting an action $a_{exec}$ via an $\epsilon$-greedy policy. Upon executing the action and observing the transition $(s_{t-1}, a_{t-1}, r_{t-1}, s_t)$, the experience is appended to $RM$. To maintain the buffer size $N$, a First-In-First-Out (FIFO) scheme is employed, where the oldest experiences are evicted to accommodate new data.

The learning phase is delayed until a "warm-up" threshold, $N_{warmup}$, is reached, ensuring that the buffer contains a sufficiently diverse distribution of environmental interactions. Once this condition is met, the algorithm performs $m$ learning iterations per environmental step. In each iteration, a transition $sm = (s, a, r, s')$ is sampled from $RM$. For each sampled experience, the system reconstructs the relevant internal sets:
\begin{enumerate}
    \item The match set $[M]$ corresponding to the historical state $s$.
    \item The action set $[A]$ containing classifiers from $[M]$ that advocated action $a$.
    \item The subsequent match set $[M]'$ corresponding to the resulting state $s'$.
\end{enumerate}

These sets serve as the foundation for the core learning mechanisms. The \textbf{Anticipatory Learning Process (ALP)} utilizes these sets to evaluate and refine the $C-A-E$ structures, while the \textbf{Reinforcement Learning (RL)} component updates the payoff expectations of the classifiers in $[A]$ based on the reward $r$ and the potential of $[M]'$. Finally, the \textbf{Genetic Algorithm (GA)} is invoked stochastically based on the $\Theta_{GA}$ threshold, which measures the average time since the classifiers in $[A]$ last participated in a genetic discovery process. This ensures that structural exploration is maintained throughout the replay process, allowing the agent to discover generalized rules from previously encountered transitions.

\subsection{Hindsight Experience Replay (HER)}
\label{back:her}
A significant challenge in RL is the \textit{sparse reward problem}, where an agent receives meaningful feedback only after reaching a specific, often distant, goal state $g$. In such environments, random exploration rarely yields a reward, leading to inefficient learning. Hindsight Experience Replay \cite{andrychowicz2017hindsight} addresses this by allowing the agent to "learn from failure."

In a goal-conditioned setting, if an agent fails to reach the intended goal $g$ and instead reaches state $s'$, HER re-processes the trajectory as if $s'$ was the original intention. By re-labeling the experience with this \textit{virtual goal}, the agent receives a non-zero reward signal. Formally, for a transition $(s, a, r, s')$ with a goal $g$, HER generates additional transitions where $g$ is replaced by a state visited later in the same episode. This transformation effectively turns a sparse reward problem into a dense one, allowing the agent to learn the underlying environmental dynamics regardless of whether the primary task was accomplished.

\section{ACS2HER: Anticipatory Classifier System with Hindsight Experience Replay}
\label{sec:acs2-her}

The ACS2HER model represents an integration of the Anticipatory Classifier System with the Hindsight Experience Replay mechanism. While ACS2 is proficient at building a cognitive map of the environment through latent learning, it often struggles in environments with sparse rewards where the goal is rarely reached during initial exploration. By incorporating HER, the model can learn from failed trials by re-labeling visited states as virtual goals.

\textbf{The fundamental idea behind ACS2HER is to treat every trajectory as a success for some goal}, even if it was not the original environmental goal $g$. When a trial concludes without reaching $g$, the algorithm selects $k$ additional states from the current trajectory using a strategy $S$. These states are designated as hindsight goals $g'$. The transitions are then stored in the Replay Memory ($RM$) with recalculated rewards. This ensures that the Anticipatory Learning Process (ALP) and Reinforcement Learning (RL) components receive sufficient updates to generalize the classifier population efficiently.

The integration of hindsight mechanisms introduces several key parameters to the ACS2 framework:

\begin{itemize}
    \item \textbf{$k$}: The hindsight factor, defining the number of additional goals generated for every real transition.
    \item \textbf{$S$}: The goal selection strategy. This determines how virtual goals are sampled from the trajectory (\textit{final} or \textit{future} states). By default, the goal strategy depends on the $k$ parameter. If it has a value of 1,  then the strategy, if not selected by the user, will take the value of \textit{final};the additional goal is always the last state achieved in the episode. For values greater than 1, the default strategy is \textit{future}; additional goals are drawn from a list of states that followed a given state. 
    \item \textbf{$m$}: The learning intensity (repeats), specifying how many transitions are sampled from the Replay Memory for update processes after each step.
    \item \textbf{$RM$}: The Replay Memory buffer, which stores both original and hindsight-enhanced experiences to break temporal correlations during learning.
\end{itemize}

The Algorithm \ref{alg:ACS2HER3} describes the logic of ACS2HER. In this version, the hindsight learning process is specifically triggered when the agent fails to reach the primary goal, focusing the computational effort on learning from mistakes.

\begin{algorithm}[ht!]
\caption{ACS2HER}
\label{alg:ACS2HER3}
\begin{algorithmic}
\Require $trials \gets$ Total number of trials
\Require $S \gets$ Chosen strategy for generating additional goals
\Require $m \gets$ Number of learning repeats per update
\Require $k \gets$ Number of additional hindsight goals
\State $current\_trial \gets 1$
\State $t \gets 0$
\State $RM \gets \emptyset$ \Comment{RM: Replay Memory buffer}
\While{$current\_trial \leq trials$}
    \State $done \gets False$
    \State $s_t \gets env.reset()$
    \State $g \gets env.goal$
    \State $trajectory \gets \emptyset$
    \While{$done$ is $False$}
        \State $a_{exec} \gets select(A)$
        \State $s_{t-1} \gets s_t$
        \State $s_t,\ r_{t},\ done \gets env.step(a_{exec})$
        \State $trajectory.add(s_{t-1},\ a_{exec},\ r_{t},\ s_{t},\ done)$
        \State $t \gets t+1$
    \EndWhile

    \If{$s_t\ \neq\ g$} \Comment{Check if the final state is NOT the target goal}
        \ForAll{$step$ in $trajectory$} \Comment{Experience buffering loop}
            \State $s_t,\ a,\ r,\ s_{t+1} \gets step$
            \State $RM.add(s_t||g,\ a,\ r_{t},\ s_{t+1}||g,\ done)$
            \State $AG \gets additional\_goals(S,\ k,\ trajectory)$
            \ForAll{$g'$ in $AG$} \Comment{HER experience buffering loop}
                \State $r' \gets reward(s_{t+1},\ g')$
                \State $RM.add(s_{t} || g',\ a,\ r',\ s_{t+1} || g',\ False)$
            \EndFor
            \State $SM$ of size $m \subseteq RM$ \Comment{Sample mini-batch from memory}
            \ForAll{$sample$ in $SM$} \Comment{Core learning loop}
                \State $s||g',\ a,\ e,\ s'||g',\ done \gets sample$
                \State $[M] \subseteq [P]$ for $s||g'$ \Comment{Identify Match Set}
                \State $[A] \subseteq [M]$ for $a$ \Comment{Identify Action Set}
                \State $[M]' \subseteq [P]$ for $s'||g'$ \Comment{Identify next Match Set}
                \State Apply ALP on $[A]$ considering $[M]',\ s||g',\ a,\ s'||g',\ t,\ done$
                \State Apply RL on $[A]$ considering $r,\ [M]'$
                \If{$do\_ga$}
                    \State Apply GA on $[A]_{t-1}$ considering $[P],\ [M]',\ s_t||g_t,\ t$
                \EndIf
            \EndFor
        \EndFor
    \EndIf
    \State $current\_trial \gets current\_trial + 1$
\EndWhile
\end{algorithmic}
\end{algorithm}

\section{Experiments}
\label{sec:experiments}

\subsection{Protocol}
ACS2 extended by Hindsight Experience Replay was implemented and evaluated in Python language on the top of pyALCS library \cite{kozlowski2018integrating}. The testing environments were created in full compliance with OpenAI Gym \cite{2016arXiv160601540B,kozlowski2018integrating}. All experiments conducted were independent and repeated 30 times, in the regime, the exploration phase $explore_{trials} = 2000$, followed by the exploitation phase $exploit_{trails} = 500$. 
The default parameters are: $\beta=0.05$, $\gamma = 0.95$, $\theta_r = 0.9$, $\theta_i=0.1$, $\epsilon = 0.5$ $\theta_{GA} = 100$, $m_u=0.3$, $\chi=0.8$, $N = 10000$, $N_{warmup} = 1000$, subsumption enabled and genetic generalization disabled. The parameters $N$ and $N_{warmup}$ are specific only for ACS2 with ER and $N$ stands for the size of the RM, $N_{warmup}$ means the length of the warm-up phase. 
All experiments were run for ACS2, ACS2ER with the parameter $m \in {2, 4, 8}$, and ACS2HER with the parameters $m \in {2, 4, 8}$ and $k \in {2, 3, 4}$.

\subsection{Metrics}
Four metrics were selected to evaluate the performance of the compared models in the environment:
\begin{itemize}
    \item \textbf{Population numerosity and reliable classifiers} (classifier $cl$ is considered \textit{reliable} when it is quality $cl.q > \theta_r$).
    \item \textbf{Knowledge} - represented by the percent of possible environment transitions for which a reliable classifier that predicts the next state successfully exists (measured only for multi-step problems).
    \item \textbf{Steps to the goal} in a trail.
    \item \textbf{Episodes in which the goal was achieved} (only for \texttt{FrozenLake} environment).
\end{itemize}

\subsection{Environments}

To evaluate the performance of the ACS2HER model, two discrete grid-world environments characterized by sparse rewards were utilized: \texttt{Maze~6} and \texttt{FrozenLake}. These environments represent typical challenges for reinforcement learning agents where the goal is rarely encountered through random exploration.

\subsubsection{\texttt{Maze~6}}
\begin{figure}[ht!]
    \centering
    \begin{tikzpicture}[scale=0.6]
        \definecolor{frozen}{RGB}{173, 216, 230} 
        \definecolor{hole}{RGB}{60, 60, 60}     
        \definecolor{goal}{RGB}{144, 238, 144}   

        \newcommand{\drawcell}[4]{
            \fill[#3] (#1,#2) rectangle (#1+1,#2+1);
            \draw[white!20, line width=0.1pt] (#1,#2) rectangle (#1+1,#2+1);
            \ifx&#4& \else \node at (#1+0.5,#2+0.5) {\bfseries #4}; \fi
        }

        \foreach \x in {0,...,8} \drawcell{\x}{8}{hole}{};

        \drawcell{0}{7}{hole}{} \drawcell{1}{7}{frozen}{} \drawcell{2}{7}{frozen}{} \drawcell{3}{7}{frozen}{} \drawcell{4}{7}{frozen}{} \drawcell{5}{7}{frozen}{} \drawcell{6}{7}{hole}{} \drawcell{7}{7}{goal}{G} \drawcell{8}{7}{hole}{}

        \drawcell{0}{6}{hole}{} \drawcell{1}{6}{frozen}{} \drawcell{2}{6}{frozen}{} \drawcell{3}{6}{hole}{} \drawcell{4}{6}{frozen}{} \drawcell{5}{6}{hole}{} \drawcell{6}{6}{hole}{} \drawcell{7}{6}{frozen}{} \drawcell{8}{6}{hole}{}

        \drawcell{0}{5}{hole}{} \drawcell{1}{5}{frozen}{} \drawcell{2}{5}{hole}{} \drawcell{3}{5}{frozen}{} \drawcell{4}{5}{frozen}{} \drawcell{5}{5}{frozen}{} \drawcell{6}{5}{frozen}{} \drawcell{7}{5}{frozen}{} \drawcell{8}{5}{hole}{}

        \drawcell{0}{4}{hole}{} \drawcell{1}{4}{frozen}{} \drawcell{2}{4}{frozen}{} \drawcell{3}{4}{frozen}{} \drawcell{4}{4}{hole}{} \drawcell{5}{4}{hole}{} \drawcell{6}{4}{frozen}{} \drawcell{7}{4}{frozen}{} \drawcell{8}{4}{hole}{}

        \drawcell{0}{3}{hole}{} \drawcell{1}{3}{frozen}{} \drawcell{2}{3}{hole}{} \drawcell{3}{3}{frozen}{} \drawcell{4}{3}{hole}{} \drawcell{5}{3}{frozen}{} \drawcell{6}{3}{frozen}{} \drawcell{7}{3}{hole}{} \drawcell{8}{3}{hole}{}

        \drawcell{0}{2}{hole}{} \drawcell{1}{2}{frozen}{} \drawcell{2}{2}{hole}{} \drawcell{3}{2}{frozen}{} \drawcell{4}{2}{frozen}{} \drawcell{5}{2}{frozen}{} \drawcell{6}{2}{frozen}{} \drawcell{7}{2}{frozen}{} \drawcell{8}{2}{hole}{}

        \drawcell{0}{1}{hole}{} \drawcell{1}{1}{frozen}{} \drawcell{2}{1}{frozen}{} \drawcell{3}{1}{frozen}{} \drawcell{4}{1}{frozen}{} \drawcell{5}{1}{frozen}{} \drawcell{6}{1}{hole}{} \drawcell{7}{1}{frozen}{} \drawcell{8}{1}{hole}{}

        \foreach \x in {0,...,8} \drawcell{\x}{0}{hole}{};

        \draw[ultra thick] (0,0) rectangle (9,9);

    \end{tikzpicture}
    \caption{The $9 \times 9$ \texttt{Maze~6} environment layout. Dark gray cells denote non-traversable walls (obstacles), light blue cells denote the traversable path, and 'G' (light green) marks the goal state.}
    \label{fig:maze6_consistent_colors}
\end{figure}
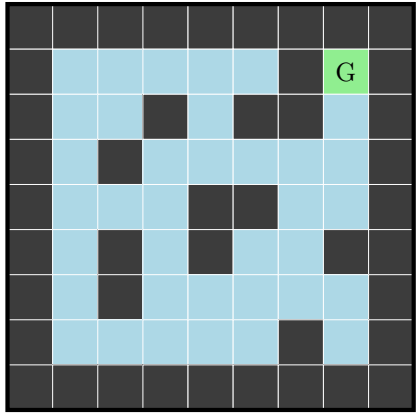

\texttt{Maze~6} is a classic environment from the \texttt{Woods} family, frequently used in the evaluation of Learning Classifier Systems. It consists of a $9 \times 9$ grid containing internal walls, traversable paths, and a single goal cell, see Fig. \ref{fig:maze6_consistent_colors}.

\begin{itemize}
    \item \textbf{State Space}: The agent perceives its environment through sensory inputs representing adjacent cells (typically 8 directions). In ACS2, these are encoded as bitstrings or symbolic strings.
    \item \textbf{Action Space}: The agent can move in 8 possible directions (N, S, E, W, and diagonals).
    \item \textbf{Reward Structure}: The environment provides a reward (e.g., 1000) only upon reaching the goal state. All other transitions result in a reward of 0.
\end{itemize}

\subsubsection{\texttt{FrozenLake}}
\texttt{FrozenLake} is a benchmark environment from the OpenAI Gym (Gymnasium) library. It simulates an agent walking across a frozen lake to reach a goal without falling into holes.

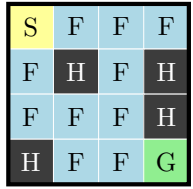
\begin{figure}[ht!]
    \centering
    \begin{tikzpicture}[scale=0.6]
        \definecolor{frozen}{RGB}{173, 216, 230} 
        \definecolor{hole}{RGB}{60, 60, 60}     
        \definecolor{goal}{RGB}{144, 238, 144}   
        \definecolor{start}{RGB}{255, 255, 153}  

        \newcommand{\drawcell}[4]{
            \fill[#3] (#1,#2) rectangle (#1+1,#2+1);
            \draw[white!20, line width=0.1pt] (#1,#2) rectangle (#1+1,#2+1);
            \ifx&#4& \else \node at (#1+0.5,#2+0.5) {\bfseries #4}; \fi
        }

        \drawcell{0}{3}{start}{S} \drawcell{1}{3}{frozen}{F} \drawcell{2}{3}{frozen}{F} \drawcell{3}{3}{frozen}{F}

        \drawcell{0}{2}{frozen}{F} \drawcell{1}{2}{hole}{\color{white}H} \drawcell{2}{2}{frozen}{F} \drawcell{3}{2}{hole}{\color{white}H}

        \drawcell{0}{1}{frozen}{F} \drawcell{1}{1}{frozen}{F} \drawcell{2}{1}{frozen}{F} \drawcell{3}{1}{hole}{\color{white}H}

        \drawcell{0}{0}{hole}{\color{white}H} \drawcell{1}{0}{frozen}{F} \drawcell{2}{0}{frozen}{F} \drawcell{3}{0}{goal}{G}

        \draw[ultra thick] (0,0) rectangle (4,4);

    \end{tikzpicture}
    \caption{The $4 \times 4$ \texttt{FrozenLake} environment layout. S: Start, F: Frozen (safe), H: Hole (failure), G: Goal (success).}
    \label{fig:frozenlake_scaled}
\end{figure}

\begin{itemize}
    \item \textbf{Grid Layout}: Typically tested on a $4 \times 4$ or $8 \times 8$ grid (in this paper, the smaller size was tested, see Fig. \ref{fig:frozenlake_scaled}). The surface consists of four types of tiles: \textit{S} (Start), \textit{F} (Frozen surface - safe), \textit{H} (Hole - episode ends in failure), and \textit{G} (Goal).
    \item \textbf{Dynamics}: The environment can be configured as "slippery." In this mode, the agent has only a partial probability of moving in the intended direction, with the remaining probability distributed among the perpendicular directions. This introduces stochasticity that tests the robustness of the ACS2's Anticipatory Learning Process.
    \item \textbf{Reward Structure}: A reward of 1 is granted only for reaching the goal \textit{G}. If the agent falls into a hole or wanders indefinitely, the reward is 0.
\end{itemize}

The combination of these two environments allows for testing the algorithm's efficiency in both deterministic, complex layouts (\texttt{Maze~6}) and stochastic, high-risk scenarios (\texttt{FrozenLake}).

\subsection{Results}
\label{sec:results}
\subsubsection{\texttt{Maze~6}}

\begin{figure}[h!tbp]
    \centering
    \includegraphics[width=\textwidth]{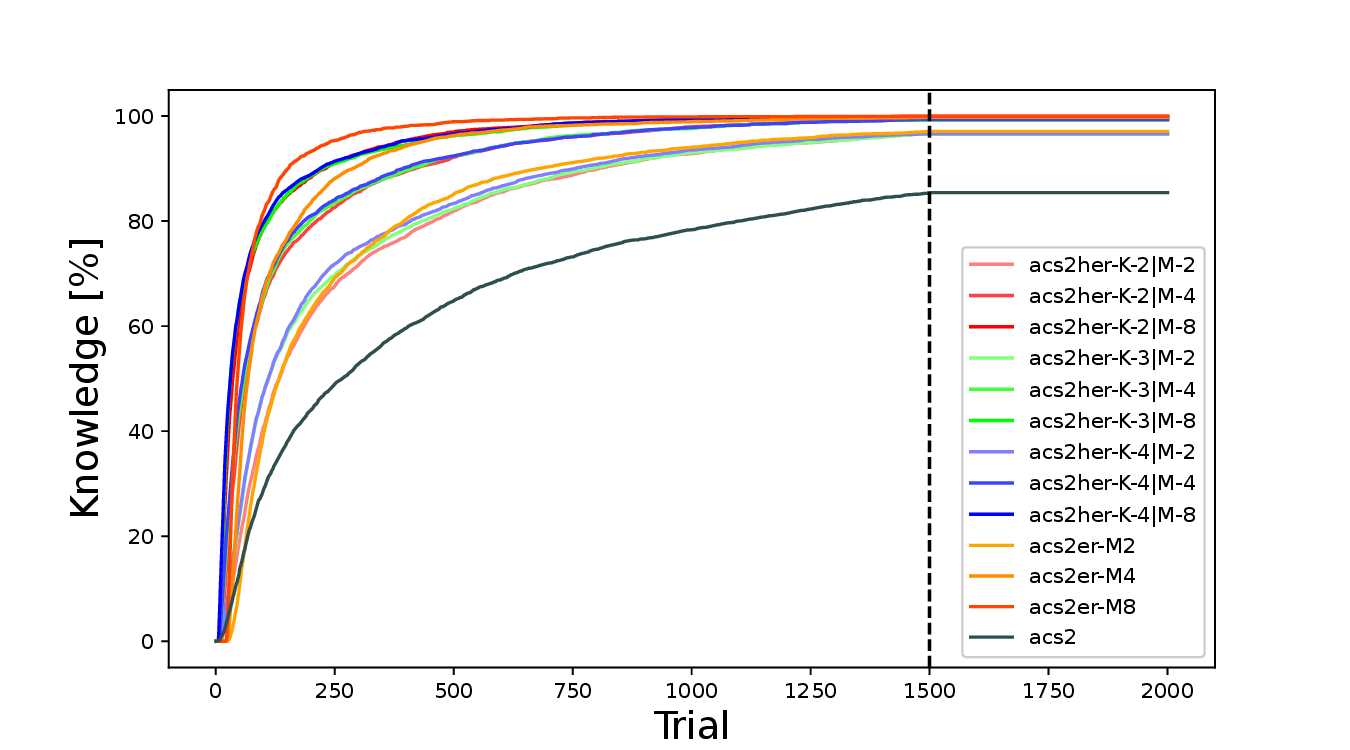}
    \caption{\texttt{Maze~6}.The knowledge}
    \label{fig:5:m6_2_knowledge}
\end{figure}

Figure \ref{fig:5:m6_2_knowledge} illustrates the percentage of environmental transitions for which a reliable classifier exists, tracking more than 2,000 trials for all models compared. \textbf{Both variants ACS2ER and ACS2HER demonstrate a significantly faster knowledge acquisition rate than the standard ACS2 model.} The results indicate that increasing the learning intensity parameter ($m$) directly accelerates the discovery of environmental dynamics.

\begin{figure}[h!tbp]
    \centering
    \includegraphics[width=\textwidth]{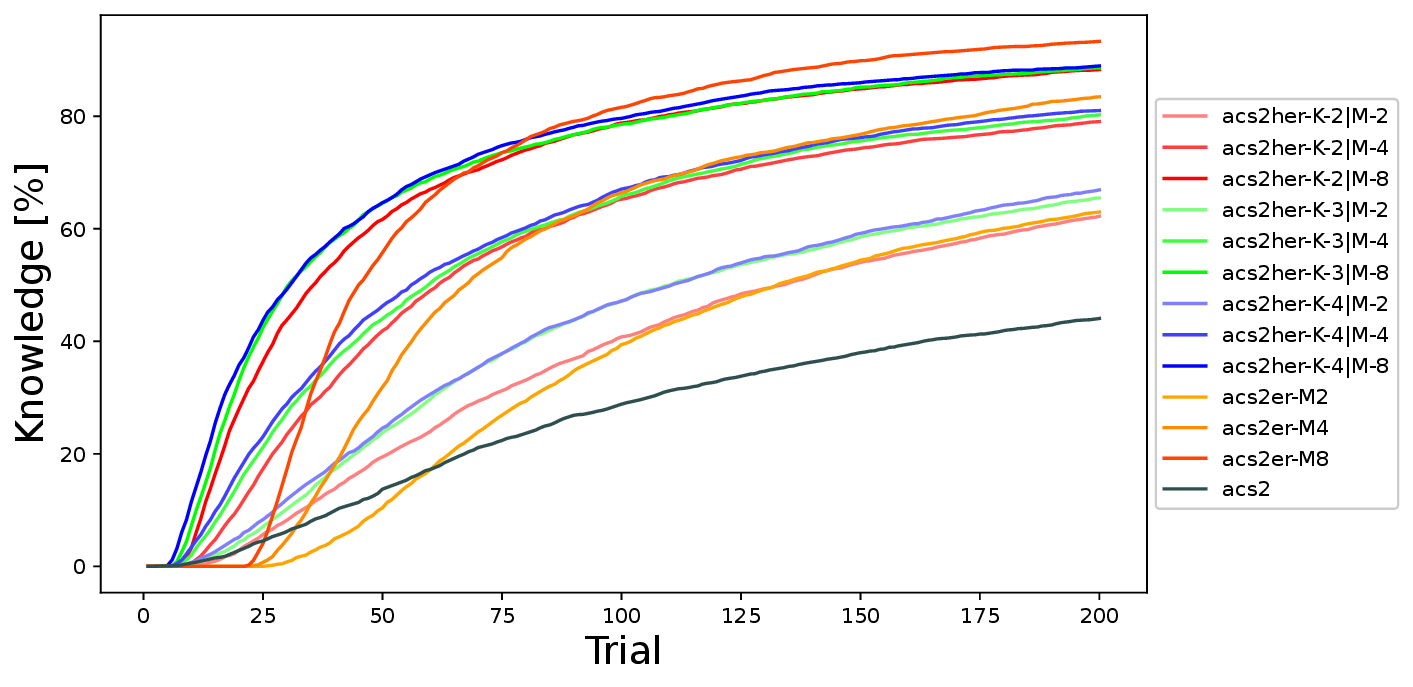}
    \caption{\texttt{Maze~6}. The knowledge up to trial 200}
    \label{fig:5:m6_2_knowledge2_200}
\end{figure}

A detailed view of the first 200 trials is also provided (Fig. \ref{fig:5:m6_2_knowledge2_200}) to highlight the immediate impact of memory-based mechanisms on the initial learning curve. The "warm-up" phase is evident; however, once learning begins, the $M=8$ variants immediately diverge from the baseline, suggesting that high-intensity replay is particularly beneficial in the earliest stages of exploration.

Table \ref{tab:m6_2_knowledge} quantifies the learning speed by identifying the specific trial number where each model reaches a knowledge threshold of 95\%, along with the maximum knowledge achieved. ACS2ER with $M=8$ is the most efficient, reaching 95\% mastery in just 239 trials, compared to 1109 trials for $M=2$. Although ACS2HER also reaches high knowledge levels (above 99\%), it generally requires more trials to reach the 95\% threshold than the equivalent configuration ACS2ER.

\begin{table}[h!tbp]
    \centering
    \caption{\texttt{Maze~6}. \textit{Average trial at 95\% knowledge achieved and the best knowledge achieved}}
    \label{tab:m6_2_knowledge}
    \begin{tabular}{|c|l|c|c|}
         \hline 
        \textbf{Model} & \textbf{Parameters} & \textbf{Trial of 95\%} & \textbf{The best knowledge [\%]} \\ \hline
        ACS2 &  & - & 85.40 \\ \hline
        ACS2ER & M-2 & 1109.0 & 97.08 \\ \hline
        ACS2ER & M-4 & 432.0 & 99.76 \\ \hline
        ACS2ER & M-8 & 239.0 & 99.95 \\ \hline
        ACS2HER & K-2 | M-2 & 1265.0 & 96.56 \\ \hline
        ACS2HER & K-2 | M-4 & 647.0 & 99.26 \\ \hline
        ACS2HER & K-2 | M-8 & 381.0 & 99.91 \\ \hline
        ACS2HER & K-3 | M-2 & 1259.0 & 96.56 \\ \hline
        ACS2HER & K-3 | M-4 & 644.0 & 99.22 \\ \hline
        ACS2HER & K-3 | M-8 & 423.0 & 99.85 \\ \hline
        ACS2HER & K-4 | M-2 & 1170.0 & 96.62 \\ \hline
        ACS2HER & K-4 | M-4 & 656.0 & 99.32 \\ \hline
        ACS2HER & K-4 | M-8 & 384.0 & 99.87 \\ \hline
    \end{tabular}
\end{table}

\begin{figure}[h!tbp]
    \centering
    \includegraphics[width=\textwidth]{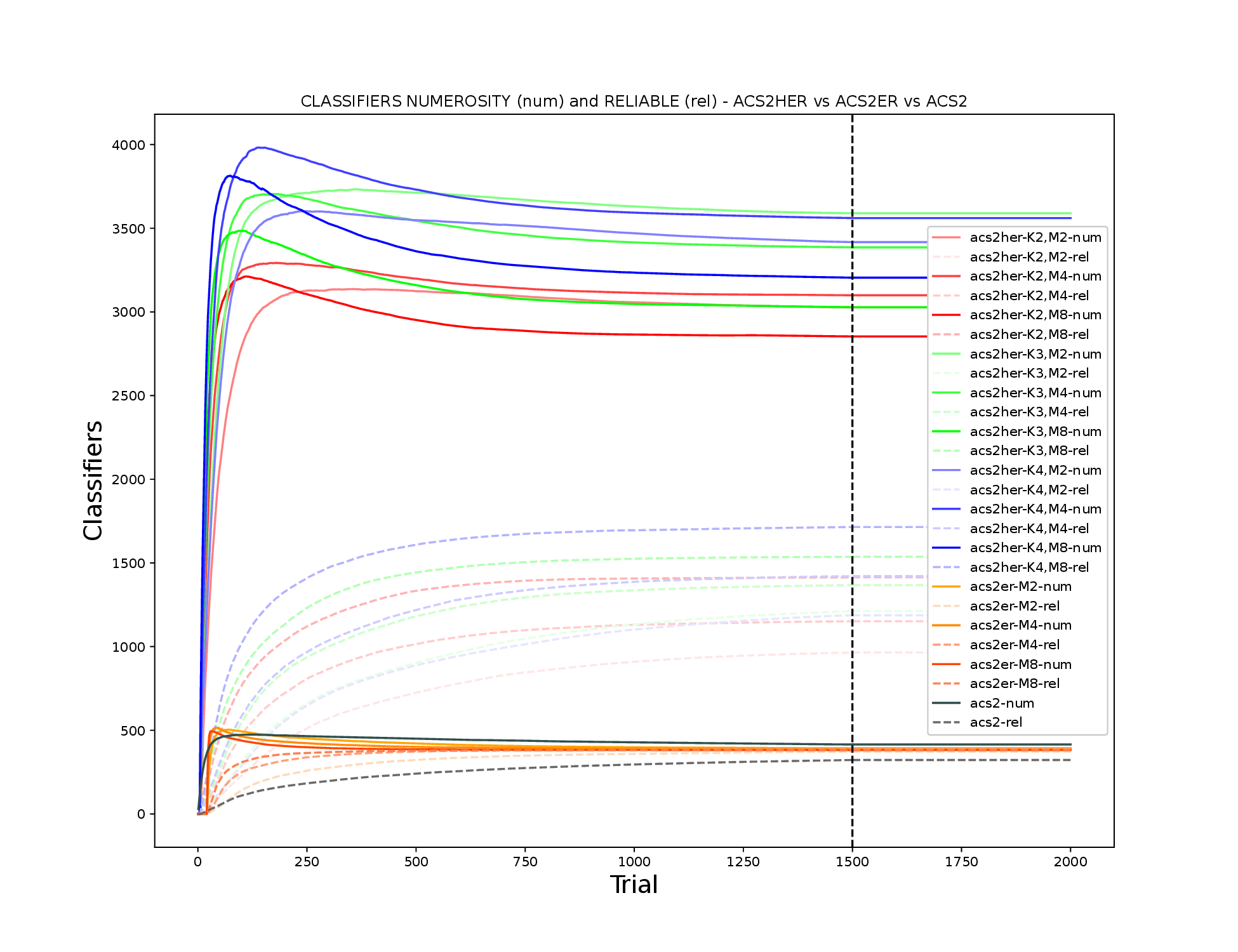}
    \caption{\texttt{Maze~6}. Numerosity of population and the number of reliable classifiers.}
    \label{fig:5:m6_2_classifiers}
\end{figure}

Figure \ref{fig:5:m6_2_classifiers} and Table \ref{tab:m6_2_classifiers} analyze the total population size (numerosity) versus the number of reliable classifiers, representing the quality and density of the agent's knowledge base.

\textbf{ACS2HER models maintain significantly larger populations} (over 3,000 classifiers) compared to ACS2 ($\sim$474). However, they also exhibit a much larger gap between total numerosity and reliable classifiers, indicating that the HER mechanism generates a high volume of experimental rules that require further validation.

\begin{table}[h!tbp]
    \centering
    \caption{\texttt{Maze~6}. Numerosity of population and the number of reliable classifiers.}
    \label{tab:m6_2_classifiers}
    \begin{tabular}{|c|l|c|c|c|c|}
        \hline
        \textbf{Model} & \textbf{Parameters} & \textbf{Max numerosity} & \textbf{Average of numerosity} & \textbf{Average of reliable} & \textbf{Difference} \\ \hline
        ACS2 & & 474.43 & 415.10 & 322.37 & 92.73 \\ \hline
        ACS2ER & M-2 & 504.17 & 392.90 & 371.67 & 21.23 \\ \hline
        ACS2ER & M-4 & 515.17 & 389.10 & 385.77 & 3.33 \\ \hline
        ACS2ER & M-8 & 495.22 & 381.78 & 381.22 & 0.56 \\ \hline
        ACS2HER & K-2 | M-2 & 3137.07 & 3027.10 & 965.23 & 2061.87 \\ \hline
        ACS2HER & K-2 | M-4 & 3293.30 & 3099.43 & 1151.67 & 1947.77 \\ \hline
        ACS2HER & K-2 | M-8 & 3212.03 & 2853.37 & 1414.17 & 1439.20 \\ \hline
        ACS2HER & K-3 | M-2 & 3733.83 & 3589.83 & 1212.23 & 2377.60 \\ \hline
        ACS2HER & K-3 | M-4 & 3704.87 & 3385.87 & 1367.40 & 2018.47 \\ \hline
        ACS2HER & K-3 | M-8 & 3485.23 & 3028.00 & 1536.83 & 1491.17 \\ \hline
        ACS2HER & K-4 | M-2 & 3602.23 & 3417.20 & 1187.30 & 2229.90 \\ \hline
        ACS2HER & K-4 | M-4 & 3982.23 & 3560.93 & 1421.10 & 2139.83 \\ \hline
        ACS2HER & K-4 | M-8 & 3813.57 & 3204.53 & 1714.83 & 1489.70 \\ \hline
    \end{tabular}
\end{table}

\begin{figure}[h!tbp]
    \centering
    \includegraphics[width=\textwidth]{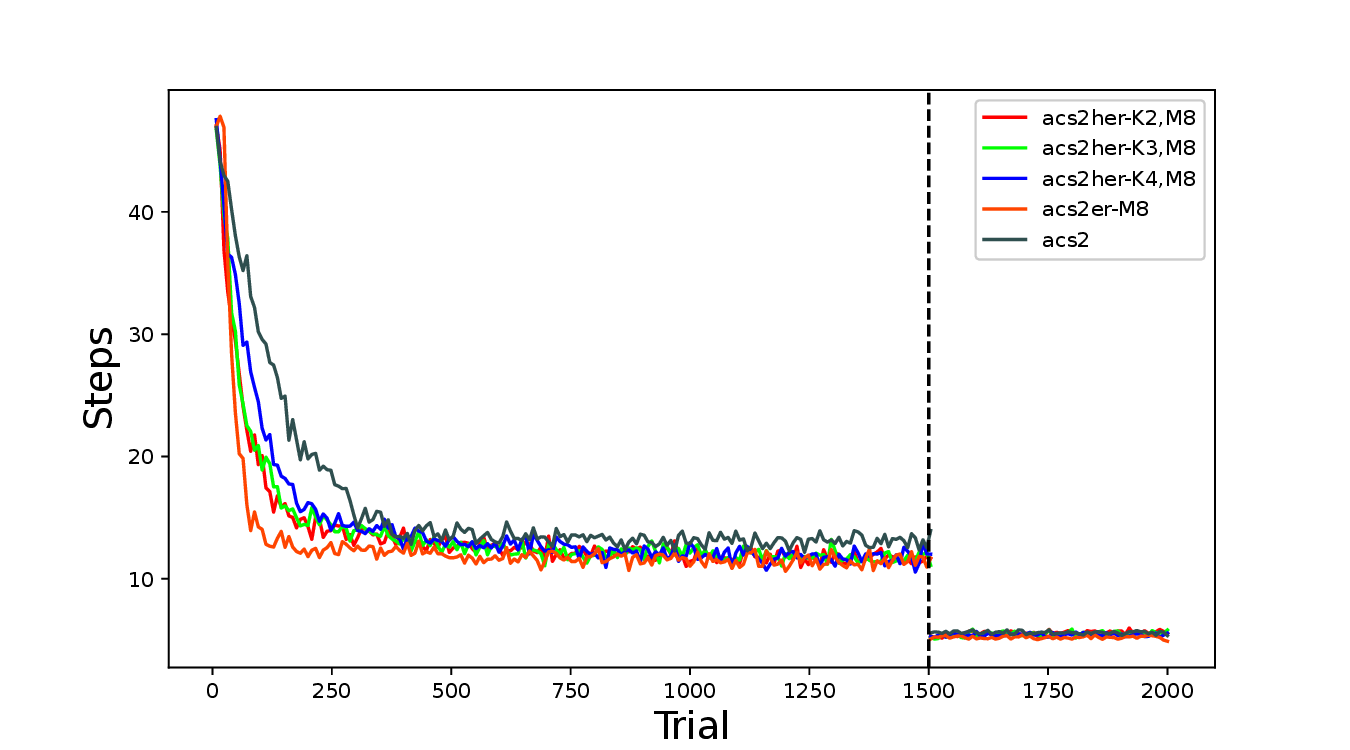}
    \caption{\texttt{Maze~6}. Average number of steps to the goal for M=8}
    \label{fig:5:m6_2_steps}
\end{figure}

Figure \ref{fig:5:m6_2_steps} and Table \ref{tab:m6_2_steps} measure the average number of steps taken to reach the goal during the exploration and exploitation phases.
All extended models eventually optimize their paths, but ACS2ER and ACS2HER achieve lower step counts more consistently than the standard ACS2. In the exploitation phase, most models converge to approximately 5.2–5.8 steps, indicating they successfully found the optimal or near-optimal path.

\begin{table}[h!tbp]
    \centering
    \caption{\texttt{Maze~6}. Average number of steps to the goal}
    \label{tab:m6_2_steps}
    \begin{tabular}{|c|l|c|c|c|}
        \hline 
        \textbf{Model} & \textbf{Parameters} & \textbf{In explore phase} & \textbf{In exploit phase} \\ \hline
        ACS2 &  & 16.01 & 5.61  \\ \hline
        ACS2ER & M-2 & 14.59 & 5.43  \\ \hline
        ACS2ER & M-4 & 13.48 & 5.26  \\ \hline
        ACS2ER & M-8 & 12.81 & 5.21  \\ \hline
        ACS2HER & K-2 | M-2 & 16.50 & 5.62  \\ \hline
        ACS2HER & K-2 | M-4 & 14.76 & 5.38  \\ \hline
        ACS2HER & K-2 | M-8 & 13.70 & 5.55  \\ \hline
        ACS2HER & K-3 | M-2 & 17.29 & 5.80  \\ \hline
        ACS2HER & K-3 | M-4 & 15.21 & 5.47  \\ \hline
        ACS2HER & K-3 | M-8 & 13.83 & 5.50  \\ \hline
        ACS2HER & K-4 | M-2 & 17.77 & 5.75  \\ \hline
        ACS2HER & K-4 | M-4 & 15.47 & 5.59  \\ \hline
        ACS2HER & K-4 | M-8 & 14.35 & 5.43  \\ \hline
    \end{tabular}
\end{table}

Table \ref{tab:m6_2_time} lists the average real-world time (in seconds) required to complete the experiment phases for each model. There is a substantial trade-off between learning efficiency and time; ACS2HER is significantly slower, taking upwards of 3,746 seconds in the explore phase compared to just 36 seconds for standard ACS2. This is due to the added complexity of generating virtual goals and the $m$-sized learning loops.

\begin{table}[h!tbp]
    \centering
    \caption{\texttt{Maze~6}. Average time to complete one experiment}
    \label{tab:m6_2_time}
    \begin{tabular}{|c|l|c|c|}
         \hline 
        \textbf{Model} & \textbf{Parameters} & \textbf{Explore phase [s]} & \textbf{Exploit phase [s]} \\ \hline
        ACS2 & & 36.30 & 3.70 \\ \hline
        ACS2ER & M-2 & 107.59 & 3.00 \\ \hline
        ACS2ER & M-4 & 164.99 & 2.82 \\ \hline
        ACS2ER & M-8 & 280.86 & 2.74 \\ \hline
        ACS2HER & K-2 | M-2 & 1190.55 & 34.47 \\ \hline
        ACS2HER & K-2 | M-4 & 1923.41 & 31.96 \\ \hline
        ACS2HER & K-2 | M-8 & 2969.21 & 28.40 \\ \hline
        ACS2HER & K-3 | M-2 & 1290.38 & 35.36 \\ \hline
        ACS2HER & K-3 | M-4 & 2207.57 & 34.85 \\ \hline
        ACS2HER & K-3 | M-8 & 3269.34 & 30.55 \\ \hline
        ACS2HER & K-4 | M-2 & 1276.91 & 33.08 \\ \hline
        ACS2HER & K-4 | M-4 & 2347.57 & 37.03 \\ \hline
        ACS2HER & K-4 | M-8 & 3746.25 & 31.78 \\ \hline
    \end{tabular}
\end{table}

\subsubsection{\texttt{FrozenLake}}

Table \ref{tab:fl_successes} records the average number of episodes in which the agent successfully reached the goal in the stochastic \texttt{FrozenLake} environment. In this specific environment, \textbf{ACS2ER (particularly with $M=10$) outperformed the standard model}, increasing successful goal reaches. Interestingly, ACS2HER variants showed lower success rates than the baseline, suggesting that in stochastic, high-risk environments like \texttt{FrozenLake}, the virtual goals generated by HER might lead to sub-optimal policy updates compared to standard experience replay.

\begin{table}[h!tbp]
    \centering
    \caption{\texttt{FrozenLake}. Average number of episodes in which the goal was achieved}
    \label{tab:fl_successes}
    \begin{tabular}{|c|l|c|c|}
    \hline
        \textbf{Model} & \textbf{Parameters} & \textbf{In explore phase (over 2000)} & \textbf{In exploit phase (over 500)} \\ \hline
        ACS2 &  & 56.13 & 46.17  \\ \hline
        ACS2ER & M-5 & 64.97 & 60.23  \\ \hline
        ACS2ER & M-10 & 67.30 & 63.30  \\ \hline
        ACS2ER & M-15 & 66.03 & 52.37  \\ \hline
        ACS2HER & K-1 | M-5 & 44.53 & 34.63  \\ \hline
        ACS2HER & K-1 | M-10 & 47.90 & 31.07  \\ \hline
        ACS2HER & K-1 | M-15 & 45.20 & 31.80  \\ \hline
        ACS2HER & K-2 | M-5 & 50.60 & 33.63  \\ \hline
        ACS2HER & K-2 | M-10 & 45.57 & 32.53  \\ \hline
        ACS2HER & K-2 | M-15 & 43.25 & 32.43  \\ \hline
        ACS2HER & K-3 | M-5 & 58.43 & 31.97  \\ \hline
        ACS2HER & K-3 | M-10 & 50.17 & 29.87  \\ \hline
        ACS2HER & K-3 | M-15 & 51.04 & 36.19  \\ \hline
    \end{tabular}
\end{table}

\begin{figure}[h!tbp]
    \centering
    \includegraphics[width=\textwidth]{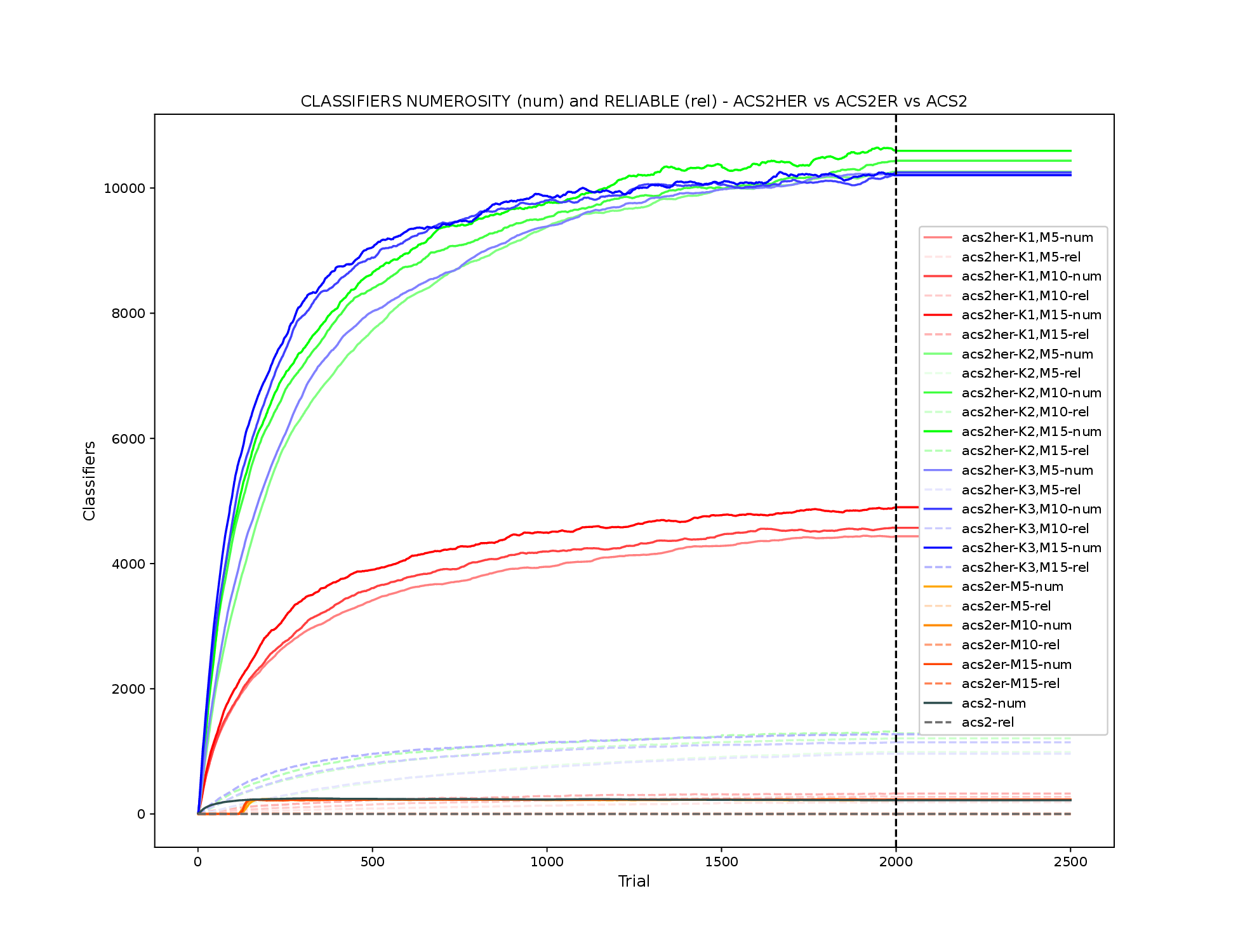}
    \caption{\texttt{FrozenLake}. Numerosity of population and the number
of reliable classifiers.}
    \label{fig:r5_fl_classifiers}
\end{figure}

Figure \ref{fig:r5_fl_classifiers} illustrates the growth of the total classifier population (numerosity) and the count of reliable classifiers over 2,500 trials in the \texttt{FrozenLake} environment. Similar to the trends observed in \texttt{Maze~6}, \textbf{ACS2HER variants produce a significantly larger number of classifiers compared to the standard ACS2 and ACS2ER models}. The substantial gap between total numerosity and reliable classifiers indicates that the hindsight mechanism generates a vast amount of experimental knowledge, much of which does not immediately meet the reliability threshold ($\theta_r$) in a stochastic grid.

Table \ref{tab:fl_steps} quantifies the efficiency of the agent's pathfinding by recording the average steps taken to reach the goal during both exploration (2,000 trials) and exploitation (500 trials). In the exploration phase, ACS2HER configurations (particularly those with $K=1$) achieved a lower average step count than the baseline ACS2. However, the data reveals that in the exploitation phase, the average steps for most models increased slightly, reflecting the difficulty of maintaining an optimal policy in the "slippery" stochastic dynamics of \texttt{FrozenLake}.

\begin{table}[h!tbp]
    \centering
    \caption{\texttt{FrozenLake}. Average number of steps to goal}
    \label{tab:fl_steps}
    \begin{tabular}{|c|l|c|c|}
    \hline
        \textbf{Model} & \textbf{Parameters} & \textbf{In explore phase} & \textbf{In exploit phase} \\ \hline
        ACS2 &  & 8.00 & 10.15 \\ \hline
        ACS2ER & M-5 & 8.40 & 12.18 \\ \hline
        ACS2ER & M-10 & 8.45 & 11.83 \\ \hline
        ACS2ER & M-15 & 8.50 & 11.05 \\ \hline
        ACS2HER & K-1 | M-5 & 7.18 & 9.72 \\ \hline
        ACS2HER & K-1 | M-10 & 6.99 & 9.20 \\ \hline
        ACS2HER & K-1 | M-15 & 7.05 & 9.44 \\ \hline
        ACS2HER & K-2 | M-5 & 7.66 & 10.04 \\ \hline
        ACS2HER & K-2 | M-10 & 7.18 & 9.81 \\ \hline
        ACS2HER & K-2 | M-15 & 7.27 & 9.28 \\ \hline
        ACS2HER & K-3 | M-5 & 8.12 & 9.28 \\ \hline
        ACS2HER & K-3 | M-10 & 7.52 & 10.03 \\ \hline
        ACS2HER & K-3 | M-15 & 7.46 & 10.33 \\ \hline
    \end{tabular}
\end{table}

\begin{figure}[h!tbp]
    \centering
    \includegraphics[width=\textwidth]{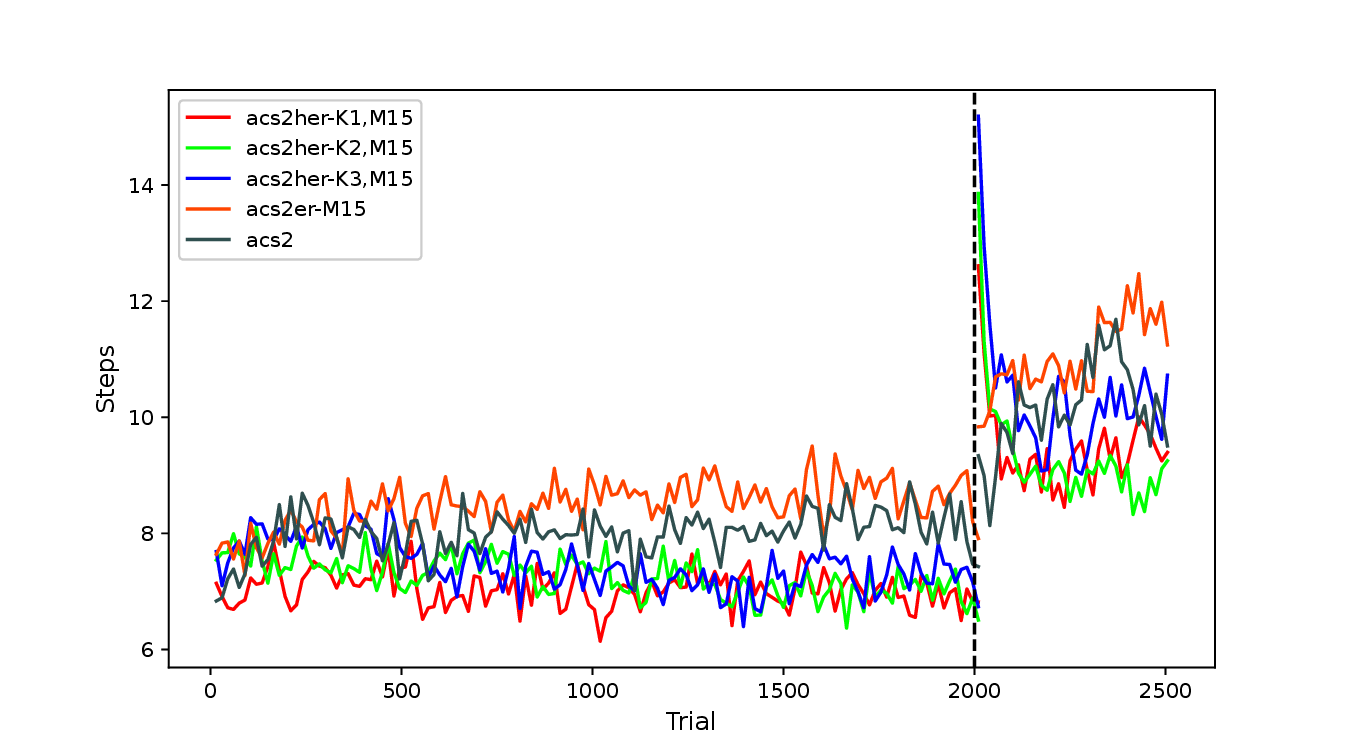}
    \caption{\texttt{FrozenLake}. Average number of steps to goal for M=15}
    \label{fig:r5_fl_steps}
\end{figure}

Figure \ref{fig:r5_fl_steps} tracks the reduction in steps to the goal over time, specifically focusing on models with the highest learning intensity parameter ($M=15$). The visualization confirms that \textbf{incorporating experience replay and hindsight goals allows the agent to find shorter paths more quickly during the early stages of learning compared to the standard ACS2 model}. The enhanced models exhibit a more rapid descent in the "steps to goal" metric during the exploration trials.

Table \ref{tab:fl_time} presents the average wall-clock time required to complete the experimental phases for each model configuration. The computational burden of ACS2HER is most pronounced in this environment, with high-intensity configurations ($K=3, M=15$) requiring over 22,491 seconds—more than 6 hours—to complete the exploration phase. Compared to the 13.06 seconds required by the standard ACS2, these results highlight that while \textbf{HER densifies learning signals, it does so at a massive cost to execution speed}.

\begin{table}[h!tbp]
    \centering
    \caption{\texttt{FrozenLake}. Average time to complete one experiment}
    \label{tab:fl_time}
    \begin{tabular}{|c|l|c|c|}
    \hline
        \textbf{Model} & \textbf{Parameters} & \textbf{Explore time [s])} & \textbf{Exploit time [s]} \\ \hline
        ACS2 & & 13.06 & 3.08 \\ \hline
        ACS2ER & M-5 & 92.73 & 3.68 \\ \hline
        ACS2ER & M-10 & 177.83 & 3.64 \\ \hline
        ACS2ER & M-15 & 262.93 & 3.37 \\ \hline
        ACS2HER & K-1 | M-5 & 3183.41 & 116.39 \\ \hline
        ACS2HER & K-1 | M-10 & 6098.30 & 114.87 \\ \hline
        ACS2HER & K-1 | M-15 & 9660.29 & 123.44 \\ \hline
        ACS2HER & K-2 | M-5 & 8415.50 & 293.52 \\ \hline
        ACS2HER & K-2 | M-10 & 14614.79 & 273.33 \\ \hline
        ACS2HER & K-2 | M-15 & 20560.25 & 249.07 \\ \hline
        ACS2HER & K-3 | M-5 & 8954.59 & 265.30 \\ \hline
        ACS2HER & K-3 | M-10 & 15538.05 & 272.29 \\ \hline
        ACS2HER & K-3 | M-15 & 22491.28 & 269.01 \\ \hline 
    \end{tabular}
\end{table}

\section{Conclusions}
\label{sec:conclusions}
This paper presented \textbf{the first integration of Hindsight Experience Replay into the Anticipatory Classifier System framework}. The experimental results across deterministic \texttt{Maze~6} and stochastic \texttt{FrozenLake} environments provide several key insights into the behavior of the ACS2HER model.

First, the inclusion of \textbf{experience replay} mechanisms (ACS2ER and ACS2HER) \textbf{significantly accelerates the rate of knowledge acquisition}. In deterministic environments like \texttt{Maze~6}, the models reached near-total environmental mastery much faster than the standard ACS2, with the learning intensity parameter $m$ acting as a primary driver of efficiency. 

Second, \textbf{the HER mechanism proves to be a powerful tool for densifying learning signals in sparse reward scenarios}. By re-labeling failed trajectories as successful virtual goals, ACS2HER maintains a much larger and more diverse population of classifiers. However, this comes with a distinct trade-off: the model generates a high volume of experimental rules that require a longer period to reach the reliability threshold, especially in stochastic settings.

Third, \textbf{the study highlights a significant computational challenge}. The time complexity of generating virtual goals and executing $m$ learning repeats per step results in a substantial increase in wall-clock time. Furthermore, in the stochastic \texttt{FrozenLake} environment, the standard HER approach did not outperform the baseline in terms of goal success rate, suggesting that "slippery" dynamics may introduce noise into the hindsight process that misleads the anticipatory model.

\section{Future Works}

The findings of this preliminary study open several avenues for future research to refine the ACS2HER architecture.

\begin{itemize}
    \item \textbf{Computational Optimization}: Future versions should investigate "Selective Replay" strategies. Instead of sampling $m$ experiences blindly, the agent could prioritize transitions with high prediction error or those that led to rare states, reducing the total number of required updates.
    \item \textbf{Adaptive Hindsight}: The parameter $k$ (number of virtual goals) could be made adaptive. Generating fewer goals as the agent's knowledge increases could prevent the population explosion (high numerosity) observed in the current experiments.
    \item \textbf{Stochastic Robustness}: To improve performance in environments like \texttt{FrozenLake}, the ALP (Anticipatory Learning Process) within the HER loop needs to be adapted to handle probabilistic transitions more effectively, perhaps by incorporating a confidence-weighted update for virtual goals.
    \item \textbf{Scaling to Continuous Spaces}: Extending ACS2HER to work with Attribute-based LCS or continuous state representations would allow for testing on more complex benchmark problems beyond simple grid worlds.
\end{itemize}

\section{Limitations}
\label{sec:limit}
Despite the promising improvements in learning speed and knowledge acquisition shown by the ACS2HER model, several limitations must be acknowledged regarding the current study and the algorithm's architecture.

\subsection{Limited Experimental Bed}
The current evaluation is restricted to two discrete grid-world environments: \texttt{Maze~6} and \texttt{FrozenLake}. Although these are standard benchmarks for Learning Classifier Systems, they represent relatively small-scale state spaces. The performance of ACS2HER in high-dimensional environments or tasks requiring continuous state-action representations remains unknown. Furthermore, the lack of a broader diversity of "Woods" family environments (e.g., \texttt{Maze 7}, \texttt{Woods 14}) prevents a full understanding of how the model scales with environmental complexity.

\subsection{Constrained Strategy Exploration}
The algorithm relies heavily on the strategy $S$ for additional goal generation and the hindsight factor $k$. In this study, only a subset of possible strategies (mainly focused on the \textit{final} state reached) was evaluated. Alternative HER strategies, such as \textit{future} (sampling goals from states visited after a specific transition) or \textit{random} (sampling goals from the entire episode), were not exhaustively compared. Additionally, the current version logic, which triggers the hindsight mechanism only upon failure ($s_t \neq g$), may be too restrictive, potentially ignoring valuable learning opportunities during successful trials.

\subsection{Statistical Robustness}
Although initial statistical observations were made, the study lacks a comprehensive suite of non-parametric statistical tests (such as the Wilcoxon signed-rank test or Friedman test with post-hoc analysis) across all performance metrics. The results are currently based on a limited number of experimental runs (seeds), which may not fully account for the high variance often observed in stochastic environments like \texttt{FrozenLake}. A more rigorous statistical verification is required to confirm that the observed gains in knowledge mastery are consistently significant.

\subsection{Computational and Architectural Overhead}
A primary bottleneck identified is the significant computational cost associated with $m$ learning repeats and the HER goal-labeling process. As shown in the results, ACS2HER can be orders of magnitude slower than standard ACS2. 
\begin{itemize}
    \item \textbf{Memory Consumption}: The Replay Memory ($RM$) and the resulting population explosion (high numerosity) place a strain on memory resources.
    \item \textbf{Rule Filtering}: The algorithm currently lacks a mechanism to prune the large volume of "hindsight-generated" rules that never reach the reliability threshold, leading to a cluttered and potentially inefficient classifier population.
\end{itemize}

\subsection{Sensitivity to Stochasticity}
The performance drop in \texttt{FrozenLake} indicates that the current implementation of HER may be overly sensitive to environmental noise. In stochastic settings, a state reached by chance might be labeled as a virtual goal, leading the Anticipatory Learning Process (ALP) to form incorrect causal links ($s, a \to s'$) that do not reflect the true underlying dynamics of the environment.

\section*{Author Contributions} 

\textbf{O.U.}: Conceptualization; Methodology; 
Formal Analysis; Validation; Writing—Original Draft; Writing—Review and Editing; Supervision; Project Administration.

\textbf{S.F.}: Methodology; Software; Investigation; Data Curation; Validation; Visualization; Writing – Review \& Editing. 

All authors reviewed and approved the final manuscript and agree to be responsible for their contributions.
All authors have read and agreed to the published version of the manuscript.

\bibliographystyle{unsrtnat}
\bibliography{bibliography} 

@String{Springer = "Springer-Verlag" }

@inproceedings{hansmeier2021experimental,
  title={An experimental comparison of explore/exploit strategies for the learning classifier system {XCS}},
  author={Hansmeier, Tim and Platzner, Marco},
  booktitle={Proceedings of the Genetic and Evolutionary Computation Conference Companion},
  pages={1639--1647},
  year={2021}
}

@article{andrychowicz2017hindsight,
  title={Hindsight experience replay},
  author={Andrychowicz, Marcin and Wolski, Filip and Ray, Alex and Schneider, Jonas and Fong, Rachel and Welinder, Peter and McGrew, Bob and Tobin, Josh and Pieter Abbeel, OpenAI and Zaremba, Wojciech},
  journal={Advances in neural information processing systems},
  volume={30},
  year={2017}
}

@article{mnih2013playing,
  title={Playing {Atari} with deep reinforcement learning},
  author={Mnih, Volodymyr and Kavukcuoglu, Koray and Silver, David and Graves, Alex and Antonoglou, Ioannis and Wierstra, Daan and Riedmiller, Martin},
  journal={arXiv preprint arXiv:1312.5602},
  year={2013}
}

@article{lin1992self,
  title={Self-improving reactive agents based on reinforcement learning, planning and teaching},
  author={Lin, Long-Ji},
  journal={Machine learning},
  volume={8},
  number={3},
  pages={293--321},
  year={1992},
  publisher={Springer}
}

@inproceedings{stein2018interpolation,
  title={What about interpolation? {A} radial basis function approach to classifier prediction modeling in {XCSF}},
  author={Stein, Anthony and Menssen, Simon and H{\"a}hner, J{\"o}rg},
  booktitle={Proceedings of the Genetic and Evolutionary Computation Conference},
  pages={537--544},
  year={2018}
}

@inproceedings{rosenbauer2020xcsf,
  title={{XCSF} with experience replay for automatic test case prioritization},
  author={Rosenbauer, Lukas and Stein, Anthony and P{\"a}tzel, David and H{\"a}hner, J{\"o}org},
  booktitle={2020 IEEE Symposium Series on Computational Intelligence (SSCI)},
  pages={1307--1314},
  year={2020},
  organization={IEEE}
}

@inproceedings{rosenbauer2020xcs,
  title={{XCS} as a reinforcement learning approach to automatic test case prioritization},
  author={Rosenbauer, Lukas and Stein, Anthony and Maier, Roland and P{\"a}tzel, David and H{\"a}hner, J{\"o}rg},
  booktitle={Proceedings of the 2020 genetic and evolutionary computation conference companion},
  pages={1798--1806},
  year={2020}
}

@inproceedings{stein2020xcs,
  title={{XCS} classifier system with experience replay},
  author={Stein, Anthony and Maier, Roland and Rosenbauer, Lukas and H{\"a}hner, J{\"o}rg},
  booktitle={Proceedings of the 2020 Genetic and Evolutionary Computation Conference},
  pages={404--413},
  year={2020}
}

@inproceedings{patzel2019survey,
  title={A survey of formal theoretical advances regarding {XCS}},
  author={P{\"a}tzel, David and Stein, Anthony and H{\"a}hner, J{\"o}rg},
  booktitle={Proceedings of the Genetic and Evolutionary Computation Conference Companion},
  pages={1295--1302},
  year={2019}
}

@InProceedings{AlgorithmicDescriptionOfACS2,
author="Butz, Martin V.
and Stolzmann, Wolfgang",
editor="Lanzi, Pier Luca
and Stolzmann, Wolfgang
and Wilson, Stewart W.",
title="An Algorithmic Description of {ACS2}",
booktitle="Advances in Learning Classifier Systems",
year="2002",
publisher="Springer Berlin Heidelberg",
address="Berlin, Heidelberg",
pages="211--229"
}

@book{hoffmann,
  title={Vorhersage und Erkenntnis},
  author={Hoffmann, Joachim},
  year={2016},
  publisher={Universit{\"a}t W{\"u}rzburg}
}

@InProceedings{IntroToACS,
author="Stolzmann, Wolfgang",
editor="Lanzi, Pier Luca
and Stolzmann, Wolfgang
and Wilson, Stewart W.",
title="An Introduction to Anticipatory Classifier Systems",
booktitle="Learning Classifier Systems",
year="2000",
publisher="Springer Berlin Heidelberg",
address="Berlin, Heidelberg",
pages="175--194"
}

@book{butz2002anticipatory,
  title={Anticipatory learning classifier systems},
  author={Butz, Martin V},
  volume={4},
  year={2002},
  publisher={Springer Science \& Business Media}
}

@article{2016arXiv160601540B,
  title={{OpenAi} {Gym}},
  author={Brockman, Greg and Cheung, Vicki and Pettersson, Ludwig and Schneider, Jonas and Schulman, John and Tang, Jie and Zaremba, Wojciech},
  journal={arXiv preprint arXiv:1606.01540},
  year={2016}
}

@incollection{holland1978cognitive,
  title={Cognitive systems based on adaptive algorithms},
  author={Holland, John H and Reitman, Judith S},
  booktitle={Pattern-directed inference systems},
  pages={313--329},
  year={1978},
  publisher={Elsevier}
}

@book{Urbanowicz:2017:ILC:3154527,
 author = {Urbanowicz, Ryan J. and Browne, Will N.},
 title = {Introduction to Learning Classifier Systems},
 year = {2017},
 edition = {1st},
 publisher = {Springer Publishing Company, Incorporated},
}

@inproceedings{kozlowski2018integrating,
  title={Integrating anticipatory classifier systems with {OpenAi Gym}},
  author={Kozlowski, Norbert and Unold, Olgierd},
  booktitle={Proceedings of the Genetic and Evolutionary Computation Conference Companion},
  pages={1410--1417},
  year={2018},
  organization={ACM}
}

@inproceedings{unold2022preliminary,
  title={Preliminary tests of an anticipatory classifier system with experience replay},
  author={Unold, Olgierd and Koz{\l}owski, Norbert and {\'S}mierzcha{\l}a, {\L}ukasz},
  booktitle={Proceedings of the Genetic and Evolutionary Computation Conference Companion},
  pages={2095--2103},
  year={2022}
}

\end{document}